%% file: main.tex
\documentclass{article}

\PassOptionsToPackage{numbers,compress}{natbib}
\usepackage[preprint]{neurips_2026}
\usepackage[utf8]{inputenc}
\usepackage[T1]{fontenc}
\usepackage{lmodern}
\usepackage{amsmath,amssymb}
\usepackage{booktabs}
\usepackage{graphicx}
\usepackage{microtype}
\usepackage{tikz}
\usepackage{pgfplots}
\usepackage{url}
\usepackage{xspace}
\usepackage{multirow}
\usepackage{xcolor}
\usepackage[hidelinks]{hyperref}
\usetikzlibrary{arrows.meta,positioning,fit,shapes.geometric,decorations.pathreplacing}
\pgfplotsset{compat=1.18}
\newcommand{\ZG}{\textsc{ZeroGraph}}

\newcommand{\ExactGraph}{\textsc{ExactGraph}}

\title{ZeroGraph: Compositional Blockwise Training via Materialized Boundary Supervision}

\author{
Omar Ramadan \\
Johns Hopkins University \\
\texttt{oramada1@jhu.edu}
\And
Sam Siavoshian \\
Independent Researcher \\
\texttt{samsiavoshian2009@gmail.com}
\And
Amir Kashif Saeed \\
Johns Hopkins University \\
\texttt{asaeed7@jhu.edu}
\AND
Benjamin A. Johnson \\
Johns Hopkins University \\
\texttt{benjamin.a.johnson@jhu.edu}
\And
Amin M. E.-A. Diab \\
Johns Hopkins University \\
\texttt{adiab3@jhu.edu}
\And
Benjamin M. Rodriguez \\
Johns Hopkins University \\
\texttt{brodrig5@jhu.edu}
}

\begin{document}
\maketitle

\input{sections/00_abstract}

\input{sections/01_introduction}
\input{sections/02_related_work}
\input{sections/03_method}
\input{sections/04_experimental_design}
\input{sections/05_results}
\input{sections/06_discussion}
\input{sections/07_conclusion}

\input{bibliography/references}
\clearpage
\appendix
\input{appendices/01_primary_qwen2_5_1_5b_systems_accounting}
\input{appendices/02_downstream_zero_shot_conditional_continuation}
\input{appendices/03_qwen2_5_3b_boundary_diagnostics}
\input{appendices/04_synthetic_and_objective_preserving_mechanics_controls}
\input{appendices/05_failure_disclosure_and_evidence_verification}

\end{document}

%% file: sections/00_abstract.tex
\begin{abstract}
Blockwise learning can remove online gradient dependencies across depth, but independently optimized blocks need not compose into a useful model. We present ZeroGraph, a systems transformation for pretrained-model conversion and distillation that materializes a frozen teacher's coherent boundary trajectory as a versioned dataset. Each partition then trains with disjoint parameters, optimizer state, random state, and checkpoints, replacing live cross-depth dependencies with persistent supervision. Experiments separate learning quality from systems behavior. On GPT-2 Large, an unanchored objective reduces denoising error by 99.44\%, yet assembled held-out cross-entropy worsens from 3.63 to 10.48 and context use becomes negative. A six-arm TinyLlama study identifies clean teacher-boundary regression as the successful interface; the tested noisy objective fails with cached or online targets. The clean recipe passes quality and context-use gates across three Qwen2.5-1.5B seeds and on a 3.086B Qwen2.5 model. In an objective-matched comparison, cached and online targets are bitwise identical. Cached jobs are faster after materialization, but a 618.31 s cache-build cost prevents wall-time break-even through 16 measured reuses. Boundary-shift tests show tolerance to intermediate representation mismatch, and an interrupted block resumes to a bitwise-identical final state without rerunning unaffected blocks. In the tested pretrained-model conversion and distillation setting, cross-depth semantic dependence can move from online optimization into persistent data.
\end{abstract}

%% file: sections/01_introduction.tex
\section{Introduction}
Distributed training systems such as FSDP, ZeRO, tensor parallelism, and pipeline parallelism reduce the cost of globally coupled optimization by redistributing model state, activation state, and computation~\cite{zero,fsdp,megatron,gpipe}. A complementary line of work asks whether different depths of a network can instead be trained from local signals~\cite{dni,greedy,diffusionblocks,blocktrain}. If successful, depth could become not only a placement boundary but also an optimization and failure boundary: partitions could own separate optimizers and checkpoints, execute on disjoint hardware, restart independently, and compose conventional sharding internally.

The central difficulty is semantic rather than mechanical. Removing inter-block gradients is straightforward once each partition is given its own local objective. The harder question is whether the resulting independently trained representation spaces remain compatible after assembly. We distinguish two properties:
\begin{equation}
\textit{blockwise trainability} \qquad \neq \qquad \textit{compositional trainability}.
\label{eq:distinction}
\end{equation}
A block is locally trainable if its own loss decreases. A collection of blocks is compositionally trainable if independently optimized partitions can be connected and retain useful end-to-end behavior.

This distinction is operationally important. On GPT-2 Large, an unanchored blockwise construction lowers all local losses, reduces assembled denoising error from 100.0182 to 0.5638, cuts peak active memory, and shortens the measured local critical path, yet the assembled language model collapses: held-out cross-entropy rises from 3.6342 to 10.4811 and all measured low-noise context effects are negative. Local success therefore does not imply compatible representation transport.

\ZG{} addresses this failure by retaining global semantic coherence while changing the form of its dependency. A frozen teacher produces boundary states \(h_0^T,\ldots,h_B^T\) on a deterministic sample stream. These states are persisted with model, data, order, shape, and content identity, forming a versioned cross-depth dataset. Block \(b\) then trains from the cached predecessor state \(h_{b-1}^T\) toward the cached successor state \(h_b^T\). The novelty claim is not hidden-state regression itself; rather, \ZG{} treats the teacher trajectory as a persistent semantic contract that converts a live depth-coupled computation into independently schedulable optimization and failure domains. The key transformation is
\begin{equation}
\boxed{\text{global semantics}}\;\longrightarrow\;\boxed{\text{persistent boundary data}}\;\longrightarrow\;\boxed{\text{independent jobs}}.
\label{eq:transformation}
\end{equation}
\ZG{} does not eliminate global information; it changes when and where that information is consumed.

This framing yields falsifiable learning and systems questions: Does a coherent boundary trajectory restore composition? Is the positive result caused by materialization or by the objective? Does caching preserve the target tensors exactly? When does materialization amortize? Do independent jobs remain independently operable under failure? Does the recipe replicate across models, seeds, and partition counts?

Our contributions are:
\begin{itemize}
    \item [] \textbf{A compositional failure result.} Mechanically independent blockwise training can satisfy local optimization and systems goals while destroying end-to-end language behavior.
    \item [] \textbf{A semantic contract for independent depth training.} Clean teacher-boundary trajectories restore composition in our setting; noisy local variants fail despite favorable teacher-forced metrics.
    \item [] \textbf{A materialized cross-depth dataflow abstraction.} We convert a globally coherent teacher computation into versioned persistent boundary data consumed by independently owned depth jobs. The contribution is relocating cross-depth semantic dependence from a live optimization path into a persistent systems interface.
    \item [] \textbf{Objective-matched systems controls.} Materialized and online clean targets are bitwise identical. Materialization removes the live teacher from cached local optimization, enables zero inter-block gradient traffic, and changes the cost structure from recurring online generation to a reusable data artifact.
    \item [] \textbf{Replication and robustness.} The clean recipe succeeds on TinyLlama 1.1B, Qwen2.5-1.5B across three seeds, and Qwen2.5-3B (3.086B parameters); it remains effective for \(B\in\{2,4,8\}\) under the measured partition sweep.
    \item [] \textbf{Independent job semantics.} A killed block resumes with optimizer, RNG, and deterministic sample-order restoration to a bitwise-identical final state, while completed neighboring blocks are not rerun.
\end{itemize}

The claim is deliberately narrower than replacing ordinary supervised fine-tuning or backpropagation. Our successful learning experiments are teacher-guided conversion and distillation of pretrained residual language models. We do not establish from-scratch global-credit-free pretraining, objective equivalence to ordinary SFT, or a universal end-to-end speedup. Accordingly, throughout the paper we treat ordinary SFT as a systems and quality reference rather than as an objective-matched algorithmic baseline.

\paragraph{Code availability.}
The accompanying ZeroGraph PyTorch package, including APIs and command-line tools for teacher-boundary materialization and independent block training, is available at \url{https://github.com/oramada/ZeroGraph}.

\begin{figure*}[t]
\centering
\includegraphics[width=0.99\textwidth]{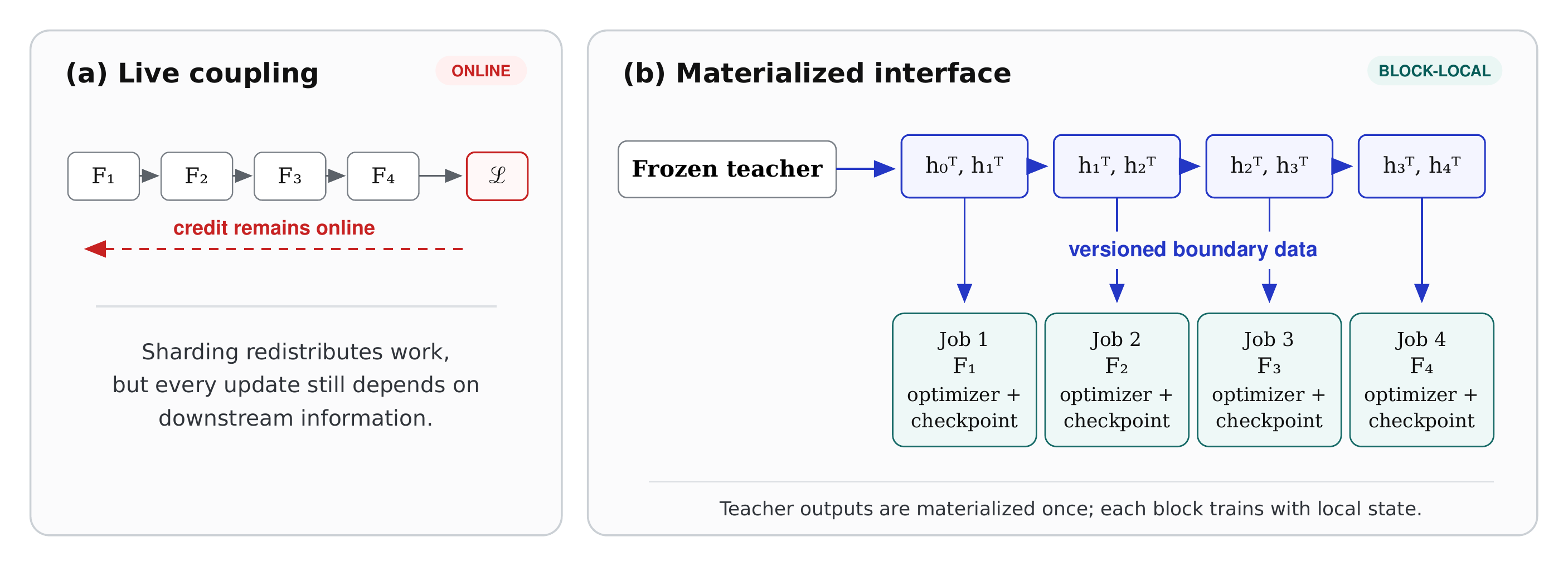}
\caption{\textbf{ZeroGraph converts a live cross-depth dependency into a persistent data interface.} Left: conventional depth-coupled optimization keeps cross-depth credit online. Right: a frozen teacher materializes a coherent trajectory once, producing versioned boundary tensors that independent block jobs consume with separate optimizer and checkpoint state.}
\label{fig:hero}
\end{figure*}

%% file: sections/02_related_work.tex
\section{Related Work}

\paragraph{Distributed training and activation management.}
Activation checkpointing and rematerialization reduce retained activations while preserving the ordinary end-to-end objective~\cite{chen2016,dtr}. ZeRO and FSDP shard parameters, gradients, and optimizer state across devices~\cite{zero,fsdp}; Megatron-style tensor parallelism and GPipe-style pipeline parallelism distribute operators or depth stages~\cite{megatron,gpipe}. These systems optimize a globally connected objective and can redistribute computation and state, but upstream updates remain dependent on downstream information. They can be applied inside a \ZG{} block, yet they do not by themselves convert depth into independent optimization and failure domains.

\paragraph{Local and blockwise learning.}
Synthetic gradients (Decoupled Neural Interfaces) and greedy layerwise learning reduce update locking by replacing or localizing downstream learning signals~\cite{dni,greedy}. DiffusionBlocks trains blocks through a diffusion interpretation of the forward process~\cite{diffusionblocks}, while BlockTrain explores decentralized local objectives and representation transport~\cite{blocktrain}. These approaches demonstrate that local signals can drive optimization, but they leave open whether the resulting representation spaces remain compatible after assembly---the compositional trainability question that is central to our work at billion-parameter scale.

\paragraph{Intermediate-state distillation.}
Knowledge distillation transfers teacher behavior to a student~\cite{hinton2015}. FitNets, Patient Knowledge Distillation, and TinyBERT match intermediate hidden representations as an auxiliary training signal~\cite{fitnets,patientkd,tinybert}. Hidden-state regression is therefore not our novelty claim; it is a well-established learning primitive. We use clean teacher-boundary regression as an experimentally identified semantic interface that restores composition, then ask a systems question: what changes when that interface is versioned, persisted as a cross-depth dataset, and consumed by independently owned depth jobs with disjoint optimizer, random, and checkpoint state?

\paragraph{Positioning.}
\ZG{} sits at the intersection of these lines. It retains a coherent global semantic source (the frozen teacher trajectory) but relocates its consumption from a live cross-depth optimization dependency into persistent boundary data. This yields independently schedulable optimization and failure domains while remaining compatible with conventional sharding inside each block. The contribution is the systems abstraction and the empirical demonstration that local trainability alone is insufficient for composition, not a new local learning rule.

%% file: sections/03_method.tex
\section{Method}

Let a residual model be partitioned into \(B\) contiguous maps
\begin{equation}
    h_b = F_b(h_{b-1};\theta_b), \qquad b=1,\ldots,B,
\label{eq:chain}
\end{equation}
with terminal task loss \(\mathcal{L}(h_B,y)\). In ordinary reverse-mode differentiation the cotangent \(\partial\mathcal{L}/\partial h_b\) is propagated through every preceding partition, so an update to \(\theta_b\) depends on downstream information. Sharding (data, tensor, or pipeline) can change where that signal is stored or computed, but does not remove the online dependence.

\subsection{Unanchored independence}
A block-local objective removes cross-depth derivatives mechanically. For a local denoising construction,
\begin{equation}
    \mathcal{L}_b(\theta_b)=\mathbb{E}\bigl[w(\sigma)\,\ell(D_b(x,z_\sigma,\sigma;\theta_b),y)\bigr],
\end{equation}
block-private interfaces imply
\begin{equation}
    \frac{\partial\mathcal{L}_b}{\partial\theta_c}=0 \qquad\text{for all }c\neq b.
\label{eq:indep}
\end{equation}
Condition~\eqref{eq:indep} establishes optimization separability only; it does not by itself establish operational independence or compatibility of learned representation spaces. We therefore distinguish \emph{blockwise trainability} (each \(\mathcal{L}_b\) decreases) from \emph{compositional trainability} (the assembled map \(F_B\circ\cdots\circ F_1\) retains useful end-to-end behavior).

\subsection{Materialized boundary supervision}
A frozen teacher \(T\) produces a deterministic trajectory \(h_0^T,\ldots,h_B^T\) on a fixed sample stream. We define the materialization operator
\begin{equation}
    \mathcal{M}(T)\;=\;\bigl\{(h_{b-1}^T,h_b^T)\;\big|\;b=1,\ldots,B\bigr\}
\label{eq:materialize}
\end{equation}
that persists each pair together with provenance (model identity, data order, shape, and content hash). Once materialized, block \(b\) trains from the cached predecessor toward the cached successor:
\begin{equation}
    \mathcal{J}_b^{\mathrm{clean}}(\theta_b)
    = \ell\!\bigl(F_b(h_{b-1}^T;\theta_b),h_b^T\bigr)+R_b(\theta_b),
\label{eq:clean}
\end{equation}
where \(R_b(\theta_b)\) denotes any declared local anchors or task terms. The pair \((h_{b-1}^T,h_b^T)\) functions as a versioned semantic contract: it supplies the cross-depth information that ordinary backpropagation would have carried online, but now as persistent data.

We also examine a noisy variant that injects diffusion-style noise into the target,
\begin{align}
    z_{b,\sigma}^T &= h_b^T+\sigma\epsilon,\\
    \mathcal{J}_b^{\mathrm{noise}}(\theta_b)
    &=\mathbb{E}\bigl[w(\sigma)\ell(D_b(h_{b-1}^T,z_{b,\sigma}^T,\sigma;\theta_b),h_b^T)\bigr]+R_b(\theta_b).
\end{align}
Materialization and objective choice are orthogonal: the same target stream can be generated online or read from storage. Empirically, only the clean contract restores composition.

\subsection{Independent job semantics and amortization}
Operational independence additionally requires disjoint trainable parameter and optimizer ownership, block-local RNG and checkpoints, immutable cached inputs, and no live autograd or execution dependency on another trainable block. The failure/restart experiment supplies separate evidence for checkpoint and failure-domain independence. Each partition owns its trainable parameters \(\theta_b\), optimizer state, RNG state, sample position, and checkpoint. Strict mode admits no trainable alias across block jobs and records zero inter-block gradient communication. Conventional sharding may still be applied inside a block. The teacher is required only during cache construction, not during subsequent cached local optimization.

Let \(T_{\mathrm{cache}}\) be cache-construction wall time and \(T_{\mathrm{asm}}\) assembly and evaluation time. For execution waves \(\mathcal{W}\), define the schedule makespan
\begin{equation}
    T_{\mathrm{sched}}=\sum_{w\in\mathcal{W}}\max_{b\in w}T_b.
\end{equation}
Then
\begin{align}
    T_{\mathrm{first}}&=T_{\mathrm{cache}}+T_{\mathrm{sched}}+T_{\mathrm{asm}},\\
    \bar{T}_{\mathrm{ZG}}(K)&=\frac{T_{\mathrm{cache}}}{K}+T_{\mathrm{sched}}+T_{\mathrm{asm}}.
\label{eq:amortized}
\end{align}
The special case \(T_{\mathrm{sched}}=\max_b T_b\) requires a simultaneous resource slot for every block. For \(B=8\) on four GPUs, two waves were required: measured four-GPU wall time was 568.923~s, whereas the single-block critical path was 266.203~s.
Equation~\eqref{eq:amortized} separates the one-time cost of relocating global semantics into persistent data from the recurring cost of independent local jobs. We therefore report post-materialization execution and cache-inclusive amortized cost separately; materialization changes the dependency structure, but is not automatically a wall-time reduction for every reuse count \(K\).

%% file: sections/04_experimental_design.tex
\section{Experimental Design}
\label{sec:experimental-design}

We organize the evidence around five research questions. For the assembled model \(F=F_B\circ\cdots\circ F_1\), define the metric vector
\[
\mathbf{M}(F)=\bigl(\mathrm{CE}(F),\mathrm{Acc}(F),\Delta_{\mathrm{ctx}}(F),G_{\mathrm{coh}}(F)\bigr).
\]
CE is minimized, while accuracy and context-use delta are maximized; the vector has no scalar ordering. Here \(\Delta_{\mathrm{ctx}}\) is the aggregate context CE delta, positive when context improves prediction. Context success additionally requires at least three of four fixed-prompt deltas to be positive. The Boolean coherence gate \(G_{\mathrm{coh}}\) passes when all four generations are nonempty, maximum token share is at most 0.12, mean unique-token fraction is at least 0.60, and repeated-bigram fraction is at most 0.25. The locked predicate is
\begin{align*}
\operatorname{Pass}(F)\iff{}&\mathrm{CE}(F)\le1.35\;\land\;\mathrm{Acc}(F)\ge0.66\\
&\land\;\Delta_{\mathrm{ctx}}(F)>0\;\land\;N_{\mathrm{prompt}+}(F)\ge3\\
&\land\;G_{\mathrm{coh}}(F),
\end{align*}
where \(N_{\mathrm{prompt}+}\) counts positive deltas among the four fixed prompts. These definitions specify the locked quality, context-use, and coherence gates. Let \(\mathcal{L}_b\) denote block \(b\)'s local objective. Boundary reconstruction error is reported as MSE.

\paragraph{RQ1: Is local trainability sufficient for composition?}
\emph{Criterion.} Blockwise trainability without compositional trainability:
\[
    \forall b:\;\mathcal{L}_b(\theta_b)\;\downarrow
    \quad\text{and}\quad
    \mathbf{M}(F)\;\text{degrades}.
\]
A GPT-2 Large matched diagnostic compares four-rank DDP with four independent nine-layer local jobs on four L4 GPUs. Both consume 1,000,192 unique tokens over 3,907 optimizer steps. We measure local losses, assembled denoising MSE, peak HBM, critical path, and end-to-end \(\mathbf{M}(F)\).

\paragraph{RQ2: What cross-depth signal restores composition?}
\emph{Criterion.} Existence of a semantic contract that restores \(\mathbf{M}(F)\) while preserving independence:
\[
    \mathcal{J}_b^{\mathrm{clean}}\;\text{succeeds},\qquad
    \mathcal{J}_b^{\mathrm{noise}}\;\text{fails}
    \quad\text{(cached or online)}.
\]
A six-arm TinyLlama 1.1B study compares ordinary FSDP2 SFT, unanchored local training, cached clean boundaries, cached noisy boundaries, online noisy boundaries, and a hierarchical two-block configuration. We record held-out CE, accuracy, free-running coherence gates, and bitwise block-state equivalence between cached and online noisy arms.

\paragraph{RQ3: Does materialization change the learning signal or only its placement?}
\emph{Criterion.} Target integrity and amortization:
\[
    \text{bitwise identity of cached vs.\ online tensors},\qquad
    \bar{T}_{\mathrm{ZG}}(K)=\frac{T_{\mathrm{cache}}}{K}+T_{\mathrm{sched}}+T_{\mathrm{asm}}.
\]
On Qwen2.5-1.5B we compare cached and online clean teacher-boundary supervision. A direct integrity test compares 12 boundaries over 32 examples and 8 steps. Sixteen paired throughput trials measure recurring wall time, GPU-hours, and cost. Cache-inclusive amortization is evaluated through \(K=16\).

\paragraph{RQ4: Does the recipe replicate?}
\emph{Criterion.} Stability of \(\mathbf{M}(F)\) across seeds and scale:
\[
    \operatorname{Pass}(F)\;\text{holds for all seeds and for a larger model}.
\]
Three independent Qwen2.5-1.5B seeds compare ordinary FSDP2 SFT with clean cached independent training. We additionally run Qwen2.5-3B (3.086B parameters) with four independently trained depth partitions, reporting CE, accuracy, context delta, and systems metrics.

\paragraph{RQ5: Are the jobs operationally independent and compositionally robust?}
\emph{Criterion.} Independent restart and tolerance to boundary mismatch:
The resumed block must reach a bitwise-identical final state, and held-out CE must remain stable under the measured predecessor interpolation and stress.

We sweep \(B\in\{2,4,8\}\), interpolate teacher and assembled predecessor states, stress observed predecessor errors up to \(4\times\), compress the boundary cache to INT8, and perform a deterministic kill/resume experiment in which one block is interrupted while completed neighboring jobs are left untouched.

Throughout, ordinary SFT is treated as a systems and quality reference rather than an objective-matched baseline, because the two procedures optimize different functionals.

%% file: sections/05_results.tex
\section{Results}

We report results against the measurement criteria defined in Section~\ref{sec:experimental-design}. Throughout, \(\mathbf{M}(F)\) denotes end-to-end compositional quality (held-out CE, accuracy, context-use delta, and the coherence gate).

\subsection{RQ1: Local trainability does not imply compositional trainability} \emph{Mapping to criterion.} We test whether \(\forall b:\;\mathcal{L}_b(\theta_b)\downarrow\) can coexist with degradation of \(\mathbf{M}(F)\).

Table~\ref{tab:gpt2} shows the GPT-2 Large failure case. The unanchored system improves every local loss window and reduces assembled denoising MSE by 99.44\%. It also reduces peak allocated HBM and the measured local critical path. Yet end-to-end language quality collapses: held-out CE rises from 3.6342 to 10.4811 and all measured low-noise context effects are negative. Thus the criterion of blockwise trainability without compositional trainability is satisfied.

\begin{table}[t]
\caption{GPT-2 Large matched diagnostic. Local optimization and systems metrics improve while the assembled language model fails.}
\label{tab:gpt2}
    \centering\small
    \begin{tabular}{@{}lrr@{}}
    \toprule
    Metric & DDP & Unanchored \ZG{} \\
    \midrule
    Peak allocated HBM & 14.71 GiB & 6.11 GiB \\
    Critical compute & 5539.30 s & 879.06 s \\
    Held-out CE & 3.6342 & 10.4811 \\
    Perplexity & 37.87 & 35633.95 \\
    Token accuracy & 34.89\% & 5.74\% \\
    Teacher-gen CE & 2.1558 & 7.5821 \\
    \bottomrule
\end{tabular}
\end{table}

\begin{figure}[t]
\centering
    \includegraphics[width=0.98\linewidth]{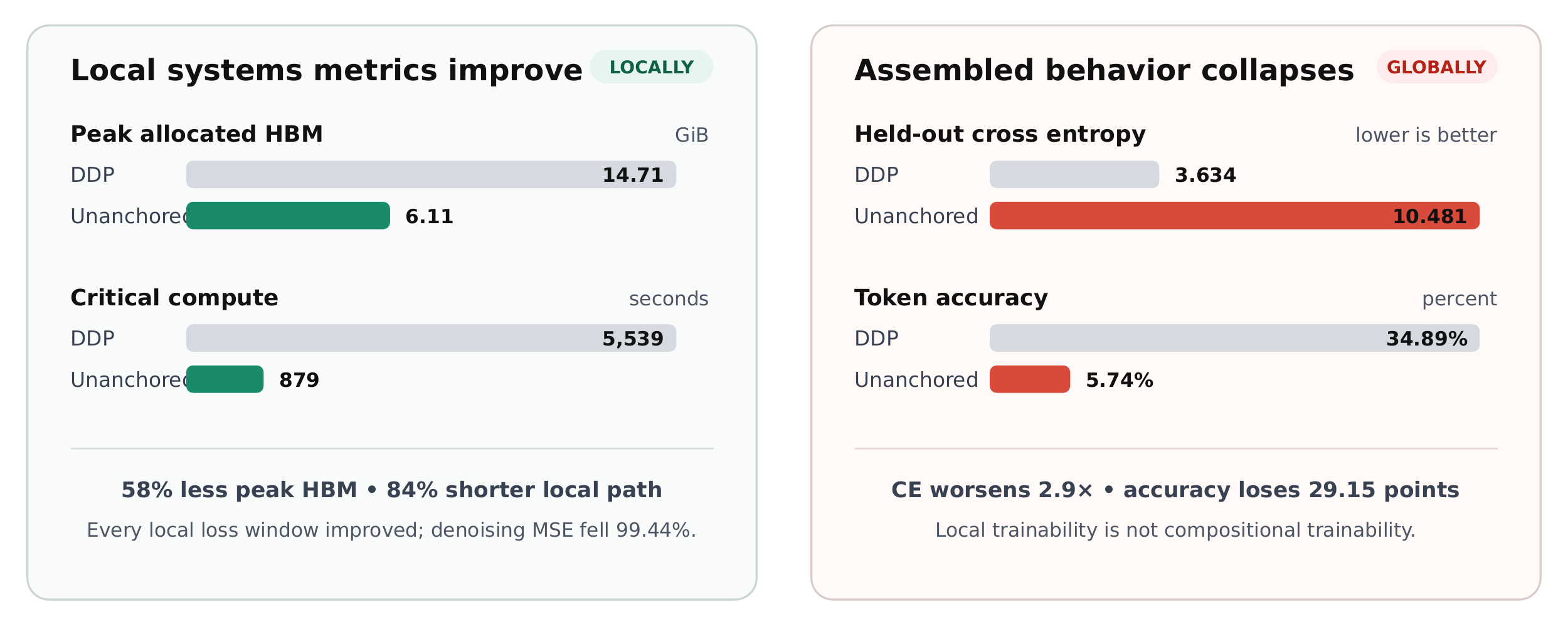}
    \caption{\textbf{Local success can coexist with global failure.} The unanchored GPT-2 Large construction lowers peak memory and the measured local critical path, but assembled held-out cross entropy rises by $2.9\times$ and token accuracy falls by 29.15 percentage points. Bar lengths are scaled within each metric; exact values are printed.}
\label{fig:local-global}
\end{figure}

All four measured low-noise context effects are negative ($-1.02\%$, $-1.03\%$, $-1.22\%$, $-2.20\%$). Figure~\ref{fig:local-global} makes the central contradiction explicit: the local systems measurements improve while assembled language behavior collapses. Neither decreasing local objectives nor low reconstruction error is sufficient evidence that independently trained blocks occupy compatible representation spaces.

\subsection{RQ2: Clean coherent boundaries provide the missing contract}
\emph{Mapping to criterion.} We test whether \(\mathcal{J}_b^{\mathrm{clean}}\) restores \(\mathbf{M}(F)\) while \(\mathcal{J}_b^{\mathrm{noise}}\) fails (cached or online).

The TinyLlama causal matrix isolates the successful interface. Cached clean boundary regression (C) reaches held-out CE 1.3059 and passes the locked context and generation gates. Adding the tested diffusion noise (D) gives similar teacher-forced CE (1.3173) but fails free-running coherence. Recomputing the same noisy targets online (E) produces bitwise-equivalent block states to D and fails identically. Thus materialization does not repair a bad local objective; the positive result is clean coherent boundary supervision, satisfying the stated criterion.

\begin{table}[t]
\caption{TinyLlama 1.1B causal study.}
\label{tab:tiny}
    \centering\scriptsize
    \begin{tabular}{@{}llrrrr@{}}
    \toprule
    Arm & Training signal & CE & Acc. & HBM & Local s \\
    \midrule
    A & FSDP2 SFT & 1.4643 & 64.78\% & 5.18 & 3564.4 \\
    B & unanchored local & 1.5554 & -- & -- & -- \\
    C & cached clean & 1.3059 & 67.57\% & 5.13 & 536.7 \\
    D & cached noisy & 1.3173 & 67.26\% & 5.13 & 531.3 \\
    E & online noisy & 1.3173 & 67.26\% & 7.18 & 578.9 \\
    F & noisy, $B=2$, inner FSDP2 & 1.3366 & 66.61\% & 4.75 & 2184.6 \\
    \bottomrule
\end{tabular}
\end{table}

\begin{figure}[t]
\centering
    \includegraphics[width=0.98\linewidth]{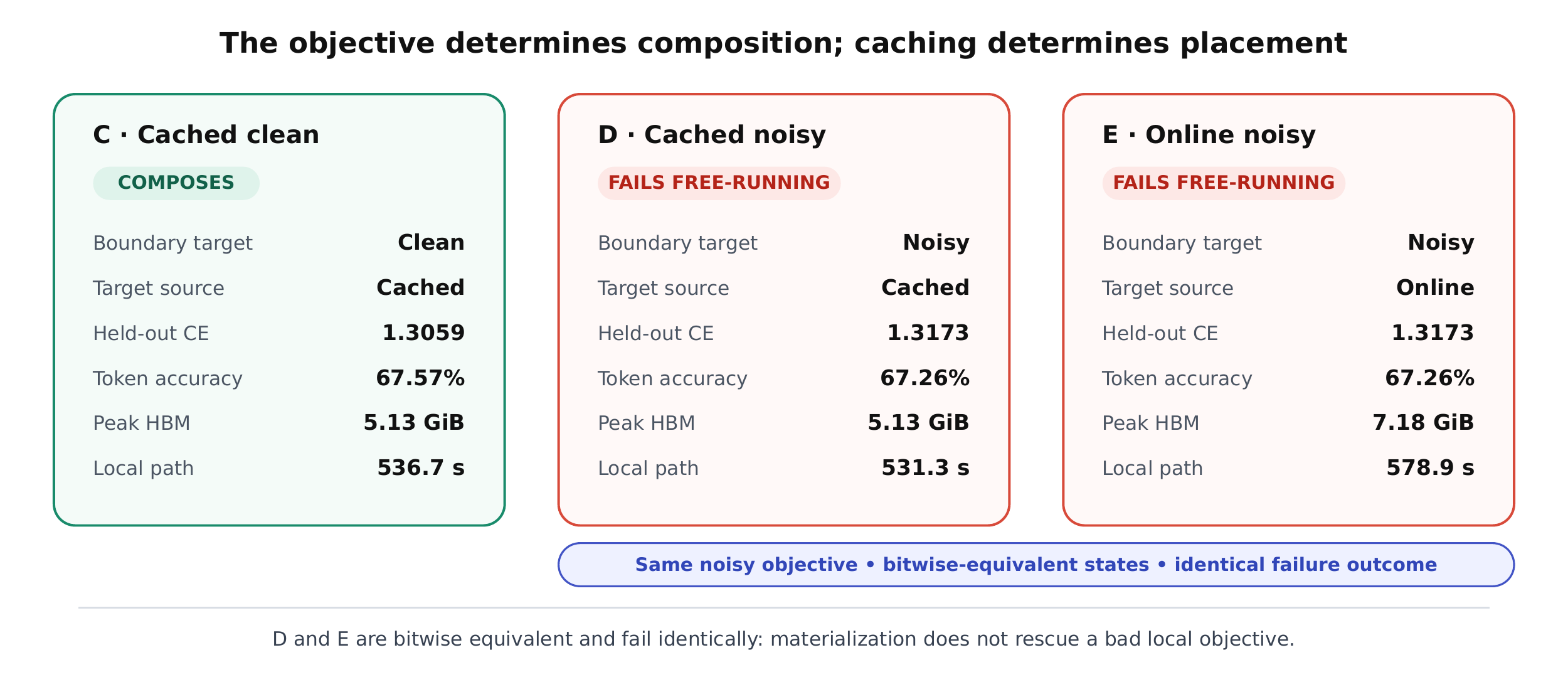}
    \caption{\textbf{The objective determines composition; caching determines placement.} Clean cached boundary regression composes. The noisy cached and noisy online arms have identical held-out CE and token accuracy, produce bitwise-equivalent block states, and fail the same free-running gates. Online target generation also raises peak HBM and local wall time.}
\label{fig:causal-matrix}
\end{figure}

Figure~\ref{fig:causal-matrix} isolates the causal conclusion. A useful local objective must not only fit its own targets, but also preserve a transportable representation interface for neighboring learned blocks.

\subsection{RQ3: Materialization changes placement, not the target tensors} \emph{Mapping to criterion.} We test bitwise identity of cached versus online tensors and evaluate the amortization formula \(\bar{T}_{\mathrm{ZG}}(K)\).

A direct target-integrity test compares 12 boundaries over 32 examples and 8 steps. Every tested boundary has bitwise-equal fraction 1.0, maximum absolute error 0, and MSE 0 between cached and online production. This establishes that the materialized interface can exactly reproduce the online clean target tensors, satisfying the integrity half of the criterion.

On the primary Qwen2.5-1.5B workload, the BF16 cache occupies 72.000~GiB and costs 618.310~s of allocation wall time to construct. After materialization, cached local jobs are faster than online clean jobs: across 16 paired trials, cached execution averages 99.129~s versus 116.386~s online. The one-time cache cost, however, dominates early reuse.

\begin{table}[t]
\caption{Objective-matched cache amortization. Wall-time break-even is not reached through the measured $K=16$.}
\label{tab:amort}
    \centering\scriptsize
    \begin{tabular}{@{}rrrrr@{}}
    \toprule
    $K$ & Cached s/run & Online s/run & Cached \$/run & Online \$/run \\
    \midrule
    1  & 723.982 & 116.353 & 0.231102 & 0.103321 \\
    2  & 412.891 & 117.168 & 0.160750 & 0.104045 \\
    4  & 256.587 & 117.050 & 0.124901 & 0.103941 \\
    8  & 176.465 & 117.166 & 0.105227 & 0.104043 \\
    16 & 137.773 & 116.386 & 0.096605 & 0.103351 \\
    \bottomrule
\end{tabular}
\end{table}

\begin{figure}[t]
\centering
    \includegraphics[width=0.98\linewidth]{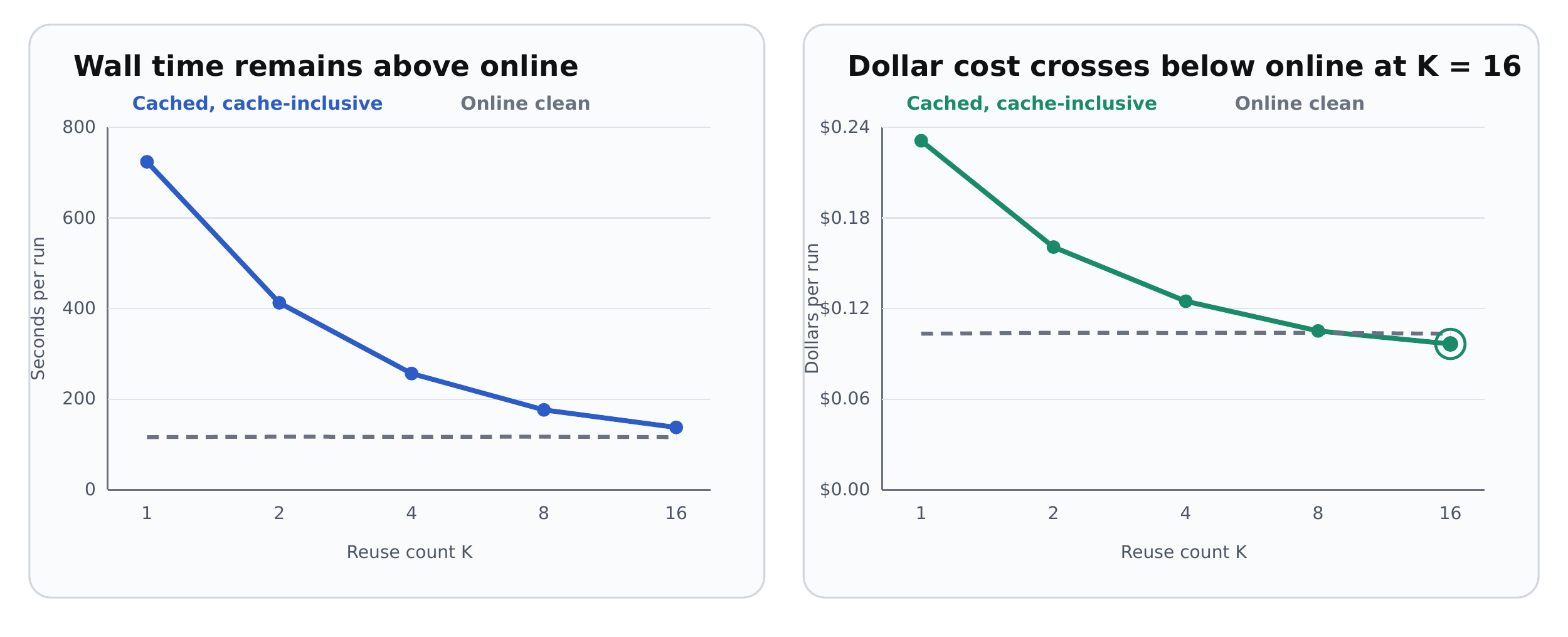}
    \caption{\textbf{Materialization creates a reuse-dependent cost regime.} Cache-inclusive wall time decreases with reuse but remains above online clean target generation through $K=16$ (left). Amortized dollar cost crosses below the online arm at $K=16$ in the measured environment (right).}
\label{fig:amortization}
\end{figure}

This is a negative result for one common interpretation of caching: materialization is not automatically a wall-time win. As Figure~\ref{fig:amortization} shows, by \(K=16\) amortized dollar cost is lower in this environment, but amortized wall time remains higher. The measured result is therefore a dependency-structure transformation with a reuse-dependent economic regime, not a universal speedup, consistent with the amortization formula.

For context, ordinary FSDP2 and ZeRO-3 SFT require 5034.264~s and 3496.523~s first-use wall time on the primary allocation, while cached clean \ZG{} requires 1169.459~s on first use and 551.149~s on reuse. These ordinary-SFT comparisons are not objective matched and are reported only as systems references. Full systems accounting appears in Appendix~\ref{app:systems}.

\subsection{RQ4: Replication across seeds and model scale} \emph{Mapping to criterion.} We test whether \(\operatorname{Pass}(F)\) holds for all seeds and for a larger model.

The Qwen2.5-1.5B clean recipe passes the locked quality gates across all three completed matched seeds (Table~\ref{tab:qwen-seeds}). The cached arm remains stable in held-out CE, accuracy, context use, and peak active HBM, satisfying the replication criterion.

\begin{table}[t]
\caption{Qwen2.5-1.5B replication. Timing is process wall for FSDP2 and four-job wall for \ZG{}. The training objectives differ, so the reported held-out CE values are evaluation metrics rather than evidence of optimization-equivalent superiority.}
\label{tab:qwen-seeds}
    \centering\scriptsize
    \begin{tabular}{@{}rrrrrr@{}}
    \toprule
    Seed & FSDP2 CE & \ZG{} CE & FSDP2 acc. & \ZG{} acc. & \ZG{} wall s \\
    \midrule
    20260910 & 1.437677 & 1.277886 & 63.970\% & 67.107\% & 546.368 \\
    20260911 & 1.438250 & 1.276192 & 63.750\% & 67.050\% & 559.248 \\
    20260912 & 1.432620 & 1.275719 & 63.778\% & 67.056\% & 569.153 \\
    \bottomrule
\end{tabular}
\end{table}

Because ordinary FSDP2 SFT and clean boundary training optimize different objectives, the lower \ZG{} held-out CE in Table~\ref{tab:qwen-seeds} should not be read as an objective-matched win over SFT. The table is instead evidence that the independently trained blocks preserve useful end-to-end behavior across repeated seeds while operating under a different teacher-guided training signal.

The same clean-boundary recipe also succeeds on Qwen2.5-3B, a 3.086B-parameter pretrained model split into four nine-layer jobs. \ZG{} achieves CE 1.300177, token accuracy 66.424\%, context delta \(+0.508852\) nat, four of four positive fixed-prompt context deltas, and teacher-scored generation CE 0.490506. The frozen base and instruct teacher have CE 1.230851 and 1.360020, respectively. All declared quality gates pass, extending the replication criterion to larger scale.

Each 3B block trains for 2,048 updates. The slowest block's compute-plus-target-acquisition path is 718.557~s; four-block wall time is 777.110~s. Peak allocated HBM is 13.594~GiB, aggregate allocated GPU-hours including I/O are 0.863456, and strict inter-block gradient traffic is zero. The five-file BF16 boundary cache occupies 10.000~GiB and takes 193.407~s allocation wall time to construct. This is a real-model scale replication, but it remains one seed and a conversion/distillation workload rather than full pretraining. Per-block boundary diagnostics for this run are given in Appendix~\ref{app:3b-diagnostics}.

\subsection{RQ5: Partition granularity, robustness, and restartability} \emph{Mapping to criterion.} We test bitwise-identical final state after kill/resume of one block, and stability of held-out CE under predecessor interpolation and stress.

\paragraph{Block-count sweep.}
Quality remains strong for \(B=2,4,8\) on Qwen2.5-1.5B. Smaller blocks reduce peak active HBM and local critical path, while \(B=8\) requires two execution waves on four GPUs and exhibits the largest maximum intermediate boundary drift.

\begin{table}[t]
\caption{Block-count sweep on Qwen2.5-1.5B.}
\label{tab:blockcount}
    \centering\scriptsize
    \begin{tabular}{@{}rrrrrr@{}}
    \toprule
    $B$ & CE & Acc. & Peak GiB & Critical s & Four-GPU wall s \\
    \midrule
    2 & 1.293729 & 66.633\% & 12.988 & 941.404 & 997.391 \\
    4 & 1.277886 & 67.107\% & 7.448 & 512.553 & 546.368 \\
    8 & 1.274216 & 67.005\% & 4.289 & 266.203 & 568.923 \\
    \bottomrule
\end{tabular}
\end{table}

\begin{figure}[t]
\centering
    \includegraphics[width=0.98\linewidth]{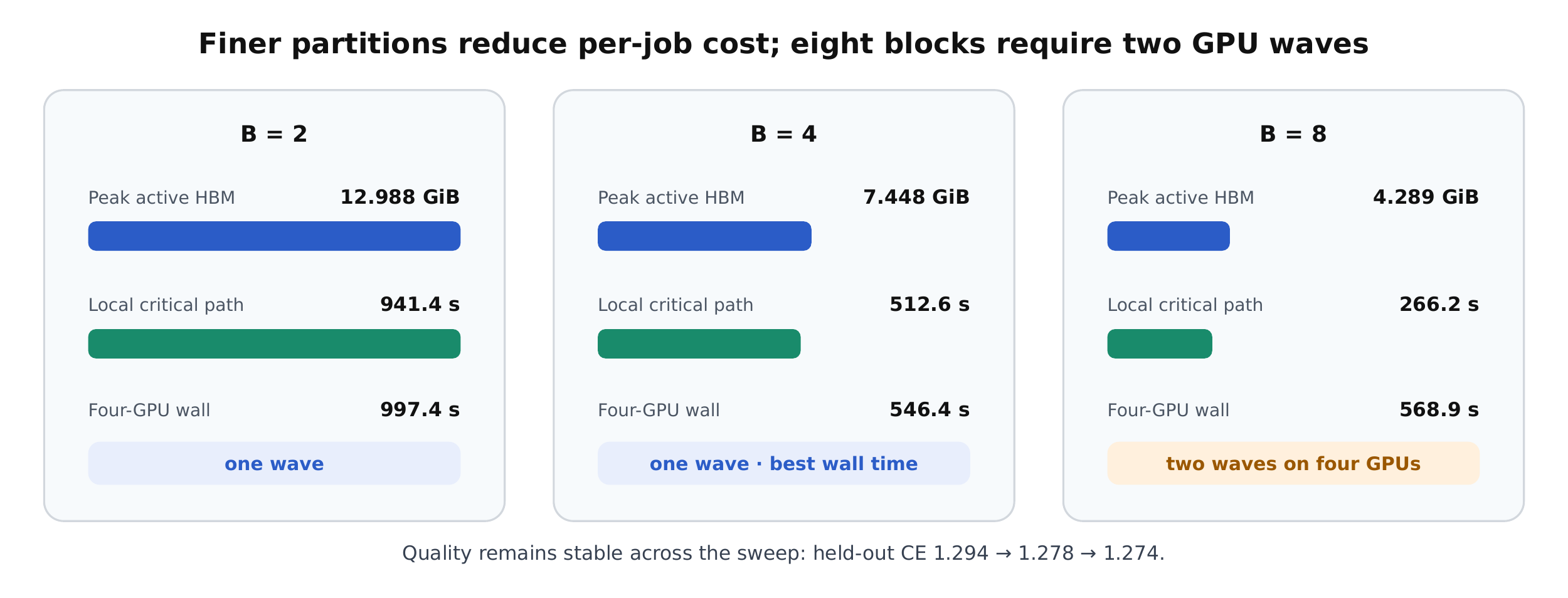}
    \caption{\textbf{Partition granularity trades per-job footprint against scheduling waves.} Increasing the partition count from two to eight reduces peak active HBM and the local critical path while preserving held-out quality. With four GPUs, however, eight blocks require two waves; the four-block configuration has the lowest measured four-GPU wall time. Bar lengths are scaled to the $B=2$ value within each metric.}
\label{fig:partition-tradeoff}
\end{figure}

Figure~\ref{fig:partition-tradeoff} visualizes the resource tradeoff behind Table~\ref{tab:blockcount}: finer partitions monotonically reduce per-job memory and local critical path, while the eight-block configuration requires two waves on four GPUs and therefore does not minimize end-to-end wall time.

\paragraph{Boundary robustness and contract tolerance.}
We interpolate between teacher predecessor states and fully assembled predecessor states to test how much deviation from the teacher interface the learned composition can tolerate. Boundary mismatch increases toward assembled predecessors, most strongly at the last block: final-block boundary MSE rises from 0.1377 under teacher input to 0.6719 under the assembled predecessor. Despite this representation drift, final CE remains in the narrow range 1.2176--1.2285 across the measured interpolation. Stressing observed predecessor error by up to \(4\times\) raises final-block boundary MSE to 3.0228 and CE to 1.2798, but still does not reproduce the catastrophic collapse of the unanchored GPT-2 run. This suggests that the learned semantic contract is not a requirement for exact pointwise state matching at inference time: in the tested regime, substantial off-teacher predecessor drift remains inside a usable compositional basin. This is an empirical tolerance result, not a guarantee for arbitrary objectives, perturbations, or architectures. Held-out CE remains within the reported range under interpolation and rises to 1.2798 under stress; these CE measurements support the robustness criterion without establishing stability of every component of the metric vector.

\paragraph{Independent restart.}
We interrupt block 1 after 256 of 512 deterministic steps, leave completed blocks 0, 2, and 3 untouched, and resume block 1 from its own checkpoint. The resumed block restores optimizer state, RNG state, and deterministic sample order, producing a final tensor state bitwise identical to the uninterrupted control. No unaffected block is rolled back or rerun. This directly validates independent failure and restart semantics (bitwise-identical final state), not merely the absence of autograd edges, satisfying the restart half of the criterion. Engineering notes on the deterministic restart certificate appear in Appendix~\ref{app:failure}.

\paragraph{Cache compression.}
Symmetric per-token INT8 values with FP16 scales reduce the 72.000~GiB BF16 cache to 36.047~GiB (1.997\(\times\) smaller). The compressed run retains essentially unchanged CE (1.277821 vs. 1.277886) and accuracy (67.152\% vs. 67.107\%) and passes the declared quality gates. The cache is lossy, so this is empirical quality retention rather than representation equivalence.

%% file: sections/06_discussion.tex
\section{Discussion}

\paragraph{What is novel?}
The successful local learning primitive is intermediate-state supervision, which has substantial prior art. \ZG{} should therefore not be read as claiming hidden-state regression as a new learning rule. The contribution is the experimentally supported systems decomposition formalized in the Method section: local optimization satisfying the independence condition~\eqref{eq:indep} is insufficient for composition; a coherent teacher trajectory can act as a versioned semantic contract via the materialization operator~\eqref{eq:materialize}; and that contract can be persisted without changing its target tensors. In this view, \ZG{} is a transformation from an online depth-coupled computation graph into an offline dataflow interface whose consumers are independently owned optimization and failure domains.

\paragraph{What does materialization buy?}
Materialization removes the live teacher and neighboring trainable blocks from the local optimization loop, enables zero inter-block gradient traffic, supports independent checkpoint/restart, and exposes the global-coherence cost as an explicit data-generation stage \(T_{\mathrm{cache}}\). It does not automatically reduce first-use wall time. In our objective-matched study the amortization formula~\eqref{eq:amortized} shows that recurring cached jobs are faster, yet the measured cache-build cost prevents wall-time break-even through 16 reuses. This is an important negative result: the architectural value and the economic value of materialization are distinct.

\paragraph{Why ordinary SFT remains a reference rather than a speed baseline.}
FSDP2 and ZeRO-3 optimize assistant-token cross entropy, whereas successful \ZG{} jobs optimize teacher boundaries plus declared local terms. Lower held-out CE or shorter training-stage time in \ZG{} therefore do not establish algorithmic superiority over ordinary SFT. The objective-matched cached-versus-online comparison is the appropriate causal systems test.

\paragraph{The teacher remains a global source of coherence.}
\ZG{} relocates global semantic information; it does not remove the need for it. Our evidence supports teacher-guided conversion and distillation of pretrained models. We do not show that useful global semantics can be created from scratch using only independent local objectives.

\paragraph{Scale and generality.}
The positive evidence spans 1.1B, 1.5B, and 3.086B pretrained language models and satisfies the replication criterion of Experimental Design. The 3B result is one controlled seed, and the larger synthetic experiment in Appendix~\ref{app:synthetic} is systems evidence only. Additional zero-shot conditional continuation results supporting semantic retention appear in Appendix~\ref{app:zeroshot}. Arbitrary PyTorch graphs, long contexts, multi-node heterogeneous clusters, and from-scratch pretraining remain open.

%% file: sections/07_conclusion.tex
\section{Conclusion}
The key failure mode in blockwise learning is not whether each partition can optimize a local objective. It is whether independently learned representation spaces remain mutually usable after assembly. Our GPT-2 result shows that these properties can diverge sharply: the independence condition can be satisfied and local losses can decrease while assembled language quality collapses.

\ZG{} resolves this failure in the tested pretrained-model setting by using a coherent teacher trajectory as a cross-depth semantic contract and materializing that trajectory, via the operator \(\mathcal{M}(T)\), as persistent boundary data. Clean boundaries restore useful assembled behavior where unanchored and noisy variants fail; cached and online targets can be bitwise identical; the recipe satisfies the replication criterion across TinyLlama, three Qwen2.5-1.5B seeds, and Qwen2.5-3B; and independent jobs can fail and resume to a bitwise-identical final state without global rollback. Materialization is not a free wall-time win---the amortization formula shows that cache construction remains significant---but it changes the dependency structure of training in a measurable way.

The resulting claim is narrow but consequential: \emph{in the tested pretrained-model conversion and distillation setting, semantic dependence across depth need not remain an online optimization dependence}. It can instead be represented as persistent cross-depth data. This separates the problem of preserving global coherence from the problem of scheduling and operating depth-wise training jobs, and provides a concrete path for studying blockwise learning as a distributed systems architecture rather than only as a local objective.

%% file: appendices/01_primary_qwen2_5_1_5b_systems_accounting.tex
\section{Primary Qwen2.5-1.5B Systems Accounting}
\label{app:systems}

The main 1.5B campaign uses Qwen/Qwen2.5-1.5B and Qwen/Qwen2.5-1.5B-Instruct on UltraChat 200k, sequence length 256, 2,048 updates, and four NVIDIA L4 GPUs where required. The model contains 1,543,714,304 parameters.

\begin{table*}[h]
\caption{Primary systems accounting. Ordinary SFT and boundary-distillation rows optimize different objectives.}
\centering\scriptsize
\setlength{\tabcolsep}{3pt}
\begin{tabular}{@{}lrrrrrrrr@{}}
\toprule
Arm & First use s & Reuse s & Critical s & Wall s & Alloc GiB & Reserved GiB & GPU-h & Ckpt GiB \\
\midrule
FSDP2 SFT & 5034.264 & 5034.264 & 5004.665 & 5031.180 & 7.473 & 11.188 & 5.560662 & 6.620 \\
ZeRO-3 SFT & 3496.523 & 3496.523 & 3480.688 & 3492.273 & 11.517 & 18.646 & 3.867346 & 2.875 \\
Global distill & 5275.348 & 4657.038 & 4628.651 & 4653.857 & 10.713 & 16.336 & 5.140341 & 6.620 \\
Cached clean $B=4$ & 1169.459 & 551.149 & 512.553 & 546.368 & 7.448 & 8.381 & 0.607075 & 4.881 \\
Online clean $B=4$ & -- & -- & 521.156 & 706.345 & 9.887 & 11.000 & 0.784828 & 4.881 \\
\bottomrule
\end{tabular}
\end{table*}

The BF16 cache contains 77,309,411,328 bytes (72.000~GiB), requires 596.688 kernel seconds and 618.310 allocation-wall seconds, consumes 0.171753 GPU-hours, and peaks at 2.950~GiB. Strict cached jobs record zero inter-block gradient traffic. Algorithmic collective payloads for ordinary FSDP2 and ZeRO-3 are 25.879~TiB/rank; global connected distillation is 21.966~TiB/rank.

%% file: appendices/02_downstream_zero_shot_conditional_continuation.tex
\section{Downstream Zero-Shot Conditional Continuation}
\label{app:zeroshot}

Each task uses 256 examples and unnormalized conditional continuation log-likelihood. These results support semantic retention, not broad benchmark superiority.

\begin{table}[h]
\caption{Mean zero-shot accuracy across ARC-Easy, BoolQ, HellaSwag, PIQA, and WinoGrande.}
\centering\small
\begin{tabular}{@{}lr@{}}
\toprule
Variant & Mean \\
\midrule
Frozen base & 65.000\% \\
Frozen teacher & 65.078\% \\
Global distillation & 63.672\% \\
Ordinary FSDP2 & 62.813\% \\
\ZG{} $B=4$ & 64.609\% \\
INT8 cached & 63.672\% \\
\bottomrule
\end{tabular}
\end{table}

%% file: appendices/03_qwen2_5_3b_boundary_diagnostics.tex
\section{Qwen2.5-3B Boundary Diagnostics}
\label{app:3b-diagnostics}

For the 3.086B run, predecessor and output mismatch increase with depth. The final block has predecessor MSE 0.03413668, teacher-input fit MSE 0.29833512, and assembled output MSE 0.80934479. Earlier blocks have assembled output MSE 0.00611112, 0.01503759, and 0.03413668. These diagnostics show accumulation of representation shift even when the final model still passes quality and context-use gates. These values are consistent with the boundary-mismatch tolerance reported for \(\mathbf{M}(F)\) in the main text.

%% file: appendices/04_synthetic_and_objective_preserving_mechanics_controls.tex
\section{Synthetic and Objective-Preserving Mechanics Controls}
\label{app:synthetic}

A 4.027B synthetic causal representation Transformer validates state ownership and working-set mechanics: one local worker peaks at 1,637,239,296 allocated bytes versus 16,133,899,776 for the ordinary control (9.854$\times$ lower), ten blocks complete 1,000 updates each, and strict inter-block gradient traffic is zero. Because the dataset and task are synthetic, this is not 4B language-quality evidence.

\ExactGraph{} provides an objective-preserving control. On a 1.007B residual Transformer, one optimizer step matches output, loss, 40 gradient tensors (2.013~GB), optimizer state, post-step parameters, inputs, and RNG bitwise. Retained non-state forward storage falls from 5,510,904 to 589,824 bytes (9.343$\times$), while peak CUDA increment remains approximately 2.014~GB because parameters and gradients dominate. The measured step is slower (70.06 vs. 59.45~ms), so this is a mechanics certificate rather than a throughput claim. ExactGraph's bitwise certificate establishes objective-preserving rematerialization under its recorded guards. It still transmits exact boundary vector--Jacobian products (VJPs), retaining the reverse dependency; it does not establish ZeroGraph's block-local or operational independence.

%% file: appendices/05_failure_disclosure_and_evidence_verification.tex
\section{Failure Disclosure and Evidence Verification}
\label{app:failure}

The deterministic restart experiment retained two failed engineering attempts. The first exposed a CUDA \texttt{map\_location} issue that restored a stored RNG ByteTensor on GPU; the loader was corrected to restore it on CPU. The second exposed nondeterministic default GPU kernels. The final certificate enabled deterministic algorithms, deterministic cuDNN, \texttt{CUBLAS\_WORKSPACE\_CONFIG=:4096:8}, math SDPA, TF32 off, and BF16 reduced-reduction off; the resumed and uninterrupted final tensor states then matched bitwise.

The full campaign reports 109 ledger entries and recorded spend of \$33.014688 against a \$50 ceiling. The authoritative evidence summary is \texttt{VERIFIED-SUMMARY.json}; the final verifier reported zero failures. Cryptographic digests belong in the released artifact manifest rather than the manuscript.

%% file: main.bbl
\begin{thebibliography}{10}
\bibitem{chen2016} T.~Chen, B.~Xu, C.~Zhang, and C.~Guestrin, ``Training deep nets with sublinear memory cost,'' in \emph{Proc. NIPS Workshop on Memory-Efficient Machine Learning}, 2016.
\bibitem{zero} S.~Rajbhandari, J.~Rasley, O.~Ruwase, and Y.~He, ``ZeRO: Memory optimizations toward training trillion parameter models,'' in \emph{Proc. SC}, 2020, pp.~1--16.
\bibitem{fsdp} Y.~Zhao, A.~Gu, R.~Varma, L.~Luo, C.-C.~Huang, M.~Xu, L.~Wright, H.~Shojanazeri, M.~Ott, S.~Shleifer, A.~Desmaison, C.~Balioglu, B.~Nguyen, G.~Chauhan, Y.~Hao, and S.~Li, ``PyTorch FSDP: Experiences on scaling fully sharded data parallel,'' \emph{Proc. VLDB Endowment}, vol.~16, no.~12, pp.~3848--3860, 2023.
\bibitem{megatron} M.~Shoeybi, M.~Patwary, R.~Puri, P.~LeGresley, J.~Casper, and B.~Catanzaro, ``Megatron-LM: Training multi-billion parameter language models using model parallelism,'' \emph{arXiv preprint arXiv:1909.08053}, 2019.
\bibitem{gpipe} Y.~Huang, Y.~Cheng, A.~Bapna, O.~Firat, M.~X.~Chen, D.~Chen, H.~Lee, J.~Ngiam, Q.~V.~Le, Y.~Wu, and Z.~Chen, ``GPipe: Efficient training of giant neural networks using model parallelism,'' in \emph{Advances in Neural Information Processing Systems}, vol.~32, 2019.
\bibitem{diffusionblocks} M.~Shing, M.~Koyama, and T.~Akiba, ``DiffusionBlocks: Block-wise neural network training via diffusion interpretation,'' in \emph{Proc. ICLR}, 2026, arXiv:2506.14202.
\bibitem{pytorch} A.~Paszke, S.~Gross, M.~Massa, A.~Lerer, J.~Bradbury, A.~Chanan, T.~Killeen, Z.~Lin, N.~Gimelshein, L.~Antiga \emph{et al.}, ``PyTorch: An imperative style, high-performance deep learning library,'' in \emph{Advances in Neural Information Processing Systems}, vol.~32, 2019.
\bibitem{dtr} M.~Kirisame, S.~Lyubomirsky, A.~Haan, J.~Brennan, M.~He, J.~Roesch, T.~Chen, and Z.~Tatlock, ``Dynamic tensor rematerialization,'' in \emph{Proc. ICLR}, 2021.
\bibitem{dni} M.~Jaderberg, W.~M.~Czarnecki, S.~Osindero, O.~Vinyals, A.~Graves, D.~Silver, and K.~Kavukcuoglu, ``Decoupled neural interfaces using synthetic gradients,'' in \emph{Proc. ICML}, 2017, pp.~1627--1635.
\bibitem{greedy} E.~Belilovsky, M.~Eickenberg, and E.~Oyallon, ``Greedy layerwise learning can scale to ImageNet,'' in \emph{Proc. AISTATS}, 2019, pp.~583--593.
\bibitem{blocktrain} P.~Toth and D.~Oprisa, ``Decentralised AI training and inference with BlockTrain,'' \emph{arXiv preprint arXiv:2606.24722}, 2026.
\bibitem{hinton2015} G.~Hinton, O.~Vinyals, and J.~Dean, ``Distilling the knowledge in a neural network,'' \emph{arXiv preprint arXiv:1503.02531}, 2015.
\bibitem{fitnets} A.~Romero, N.~Ballas, S.~E.~Kahou, A.~Chassang, C.~Gatta, and Y.~Bengio, ``FitNets: Hints for thin deep nets,'' in \emph{Proc. ICLR}, 2015.
\bibitem{patientkd} S.~Sun, Y.~Cheng, Z.~Gan, and J.~Liu, ``Patient knowledge distillation for BERT model compression,'' in \emph{Proc. EMNLP-IJCNLP}, 2019, pp.~4323--4332.
\bibitem{tinybert} X.~Jiao, Y.~Yin, L.~Shang, X.~Jiang, X.~Chen, L.~Li, F.~Wang, and Q.~Liu, ``TinyBERT: Distilling BERT for natural language understanding,'' in \emph{Findings of EMNLP}, 2020, pp.~4163--4174.
\end{thebibliography}
